\documentclass{article}

\usepackage[preprint, nonatbib]{neurips_data_2023}

\usepackage[utf8]{inputenc} 
\usepackage[T1]{fontenc}    
\usepackage{hyperref}       
\usepackage{url}            

\usepackage{booktabs}       
\usepackage{amsfonts}       
\usepackage{nicefrac}       
\usepackage{microtype}      
\usepackage{threeparttable}
\usepackage{url}
\usepackage{graphicx}
\usepackage{caption}
\usepackage{subcaption}
\usepackage{xcolor}         
\usepackage{xspace}
\usepackage{algorithm}
\usepackage{algpseudocode}
\algnewcommand\algorithmicinput{\textbf{INPUT:}}
\algnewcommand\INPUT{\item[\algorithmicinput]}
\algnewcommand\algorithmicoutput{\textbf{OUTPUT:}}
\algnewcommand\OUTPUT{\item[\algorithmicoutput]}
\algnewcommand\algorithmicparams{\textbf{PARAMETERS:}}
\algnewcommand\PARAMETERS{\item[\algorithmicparams]}

\usepackage{enumitem}
\usepackage[neverdecrease]{paralist}

\newcommand{\dataset}{\texttt{HEADLINES}\xspace}

\usepackage{xcolor, soul}
\sethlcolor{lightgray}

\title{A Massive Scale Semantic Similarity Dataset of Historical English}

\author
{Emily Silcock$^{1}$, Melissa Dell$^{1, 2^\ast}$ \\
\normalsize{$^{1}$Harvard University; Cambridge, MA, USA.}\\
\normalsize{$^{2}$National Bureau of Economic Research; Cambridge, MA, USA.}\\
\normalsize{$^\ast$Corresponding author:  melissadell@fas.harvard.edu.}
}

\begin{document}

\maketitle

\begin{abstract}
  A diversity of tasks use language models trained on semantic similarity data. While there are a variety of datasets that capture semantic similarity, they are either constructed from modern web data or are relatively small datasets created in the past decade by human annotators. This study utilizes a novel source, newly digitized articles from off-copyright, local U.S. newspapers, to assemble a massive-scale semantic similarity dataset spanning 70 years from 1920 to 1989 and containing nearly 400M positive semantic similarity pairs. Historically, around half of articles in U.S. local newspapers came from newswires like the Associated Press. While local papers reproduced articles from the newswire, they wrote their own headlines, which form abstractive summaries of the associated articles. We associate articles and their headlines by exploiting document layouts and language understanding. We then use deep neural methods to detect which articles are from the same underlying source, in the presence of substantial noise and abridgement. The headlines of reproduced articles form positive semantic similarity pairs. The resulting publicly available \texttt{HEADLINES} dataset is significantly larger than most existing semantic similarity datasets and covers a much longer span of time. It will facilitate the application of contrastively trained semantic similarity models to a variety of tasks, including the study of semantic change across space and time. 
\end{abstract}

\section{Introduction}

Transformer language models contrastively trained on large-scale semantic similarity datasets are integral to a variety of applications in natural language processing (NLP). 
Contrastive training is often motivated by the anisotropic geometry of pre-trained transformer models like BERT \cite{ethayarajh2019contextual}, which complicates working with their hidden representations. Representations of low frequency words are pushed outwards on the hypersphere, the sparsity of low frequency words violates convexity, and the distance between embeddings is correlated with lexical similarity. This leads to poor alignment between semantically similar texts and poor performance when individual term representations are pooled to create a representation for longer texts \cite{reimers2019sentence}.
Contrastive training reduces anisotropy \cite{wang2020understanding}. 

A variety of semantic similarity datasets have been used for contrastive training \cite{muennighoff2022mteb}. Many of these datasets are relatively small, and the bulk of the larger datasets are created from recent web texts; \textit{e.g.}, positive pairs are drawn from the texts in an online comment thread or from questions marked as duplicates in a forum.
To provide a semantic similarity dataset that spans a much longer length of time and a vast diversity of topics, this study develops \dataset (\textbf{H}istorical \textbf{E}normous-Scale \textbf{A}bstractive \textbf{D}up\textbf{LI}cate \textbf{Ne}ws \textbf{S}ummaries), a massive dataset containing nearly 400 million high quality semantic similarity pairs drawn from 70 years of off copyright U.S. newspapers. Historically, around half of content in the many thousands of local newspapers across the U.S. was taken from centralized sources such as the Associated Press wire \cite{guarneri2017newsprint}. Local newspapers reprinted wire articles but wrote their own headlines, which form abstractive summaries of the articles. Headlines written by different papers to describe the same wire article form positive semantic similarity pairs. 

To construct \dataset, we digitize front pages of off-copyright local newspapers, localizing and OCRing individual content regions like headlines and articles. The headlines, bylines, and article texts that form full articles span multiple bounding boxes - often arranged with complex layouts - and we associate them using a model that combines layout information and language understanding \cite{liu2019roberta}. Then, we use neural methods from \cite{silcock2023noise} to accurately predict which articles come from the same underlying source, in the presence of noise and abridgement.
\dataset allows us to leverage the collective writings of many thousands of local editors across the U.S., spanning much of the 20th century, to create a massive, high-quality semantic similarity dataset.
\dataset captures semantic similarity with minimal noise, as positive pairs summarize the same underlying texts. 

This study is organized as follows. Section \ref{sec:dataset} describes \dataset, and Section \ref{sec:lit} relates it to existing datasets. Section \ref{sec:methods} describes and evaluates the methods used for dataset construction, Section \ref{sec:benchmark} benchmarks the dataset, and Section \ref{sec:limits} discusses limitations and intended usage. 

\section{Dataset Description} \label{sec:dataset}

\begin{figure}[ht] 
    \centering
    \includegraphics[width=.6\linewidth]{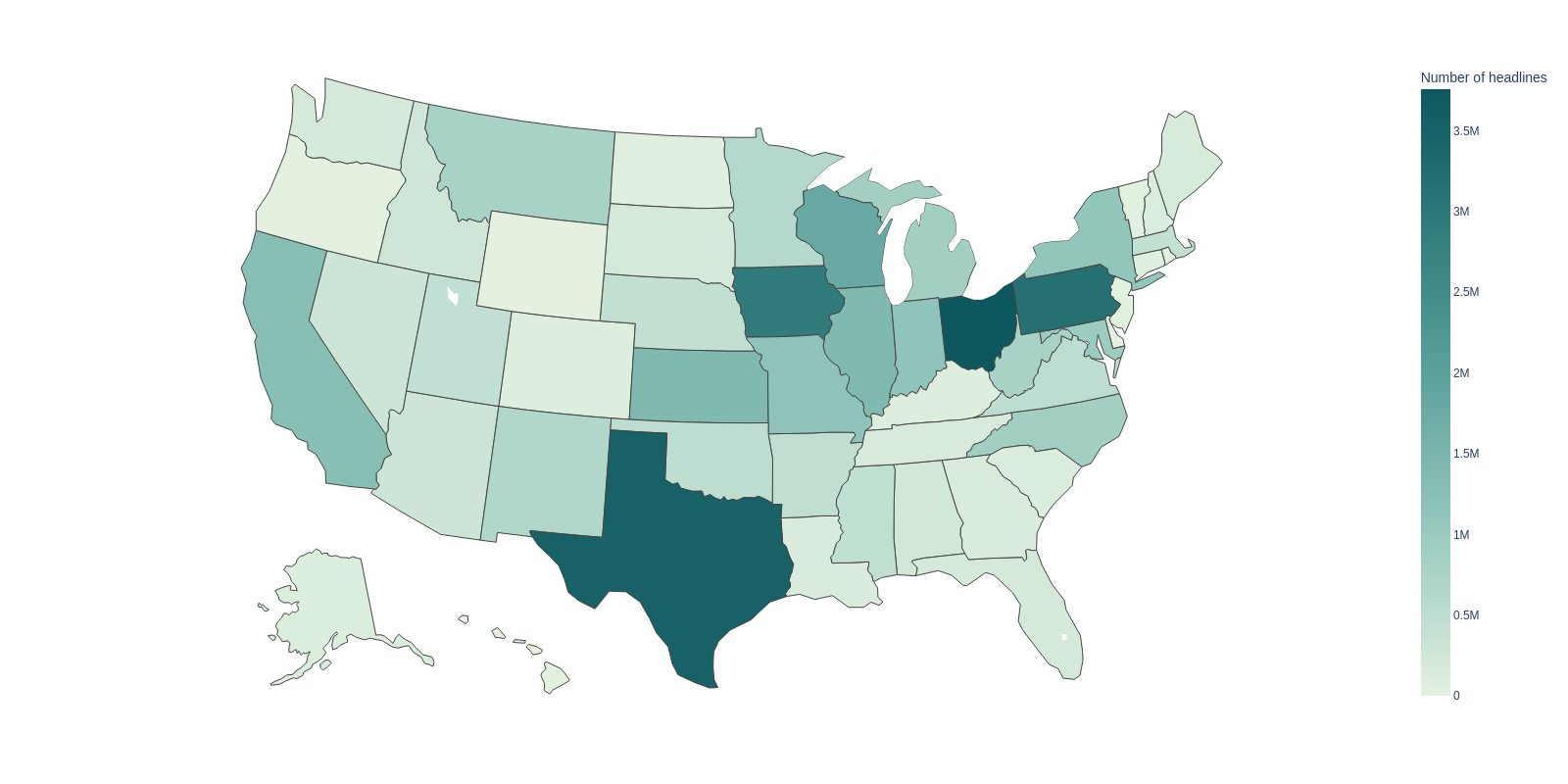}
    \caption{Geographic variation in source of headlines}
    \label{fig:map}
\end{figure}

\dataset contains 393,635,650 positive headline pairs from off-copyright newspapers. 
 Figure \ref{fig:map} plots the distribution of content by state.

Dataset statistics by decade are provided in Table \ref{tab_desc}.  Content declines sharply in the late 1970s, due to a major copyright law change effective January 1, 1978.

\begin{table}[ht]
    \centering
    \resizebox{.85\linewidth}{!}{
    \begin{threeparttable}
       \begin{tabular}{@{}ccccccccc@{}}
\toprule
Decade & Headline & Cluster & Positive & Word  & Words Per  &  Line & Lines Per & Character \\
& Count & Count & Pair Count & Count & Headline & Count & Headline & Error Rate \\\midrule
1920s & 4,889,942 & 1,032,108 & 28,928,226 & 68,486,589 & 14.0 &  18,983,014 & 3.9 & 4.3\% \\
1930s & 5,519,472 & 1,126,566 & 37,529,084 & 75,210,423 & 13.6 &  21,905,153 & 4.0 & 3.7\% \\
1940s & 6,026,940 & 1,005,342 & 62,397,004 & 61,629,003 & 10.2 &  19,538,729 & 3.2 & 2.4\% \\
1950s & 7,530,810 & 1,192,858 & 100,527,238 & 61,127,313 & 8.1 &  20,823,786 & 2.8 & 2.3\% \\
1960s & 6,533,071 & 926,819 & 108,415,279 & 46,640,311 & 7.1 &  16,408,148 & 2.5 & 3.7\% \\
1970s & 3,664,201 & 585,782 & 52,981,097 & 24,472,831 & 6.7 &  7,829,510 & 2.1 & 3.2\% \\
1980s & 703,052 & 170,507 & 2,857,722 & 5,161,537 & 7.3 &  1,502,893 & 2.1 & 1.5\% \\
\textbf{Total} & \textbf{34,867,488} & \textbf{6,039,982} & \textbf{393,635,650} & \textbf{342,728,007} & \textbf{9.8} & \textbf{106,991,233} & \textbf{3.1} \\ \bottomrule
&&&&&
    \end{tabular}
    \end{threeparttable}
  }
    \caption{Descriptive statistics of \dataset.}
      \label{tab_desc}
\end{table}

The supplementary materials summarize copyright law for works first published in the United States. The newspapers in \dataset are off-copyright because they were published without a copyright notice or did not renew their copyright, required formalities at the time. 
Far from being an oversight, it was rare historically to copyright news, outside the nation’s most widely circulated papers.
The headlines in our dataset were written by editors at these local papers, and hence are in the public domain and anyone can legally use or reference them without permission. 

It is possible that a newspaper not itself under copyright could reproduce copywritten content from some third party - the most prevalent example of this is comics - but this does not pose a problem for \dataset, since the dataset is built around the locally written headlines that describe the same wire articles. If we were to accidentally include a syndicated headline, it would be dropped by our post-processing, since we drop headline pairs within a Levenshtein edit distance threshold of each other. It is also worth noting that a detailed search of U.S. copyright catalogs by \cite{copyright} did not turn up a single instance of a wire service copyrighting their articles. (Even if they had, however, it would not pose a problem for headlines, since they were written locally.)

\begin{figure}[ht] 
    \centering
    \includegraphics[width=.6\linewidth]{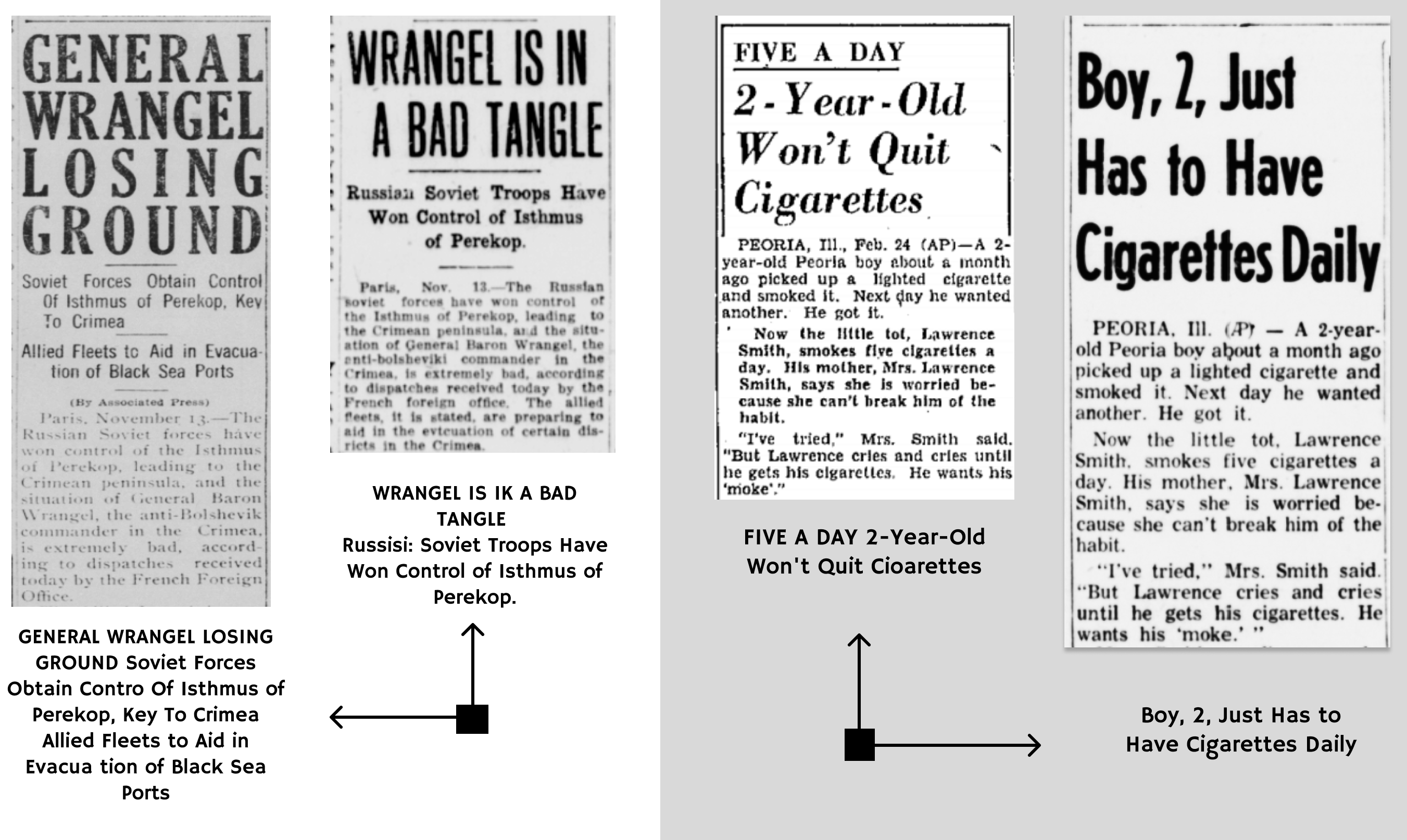}
    \caption{Semantic similarity examples, showing article image crops and OCR'ed headlines.}
    \label{fig:pairs}
\end{figure}

Figure \ref{fig:pairs} shows examples of semantic similarity pairs.

We quantify variation in \dataset across years, using a measure reminiscent of Earth Mover distance. This measure computes how much each text in a query dataset (e.g., 1920 headlines) would have to change (in embedding space) to have the same representation as the closest text in a key dataset (e.g., 1930 headlines). 

\begin{figure}[ht] 
    \centering
    \includegraphics[width=.7\linewidth]{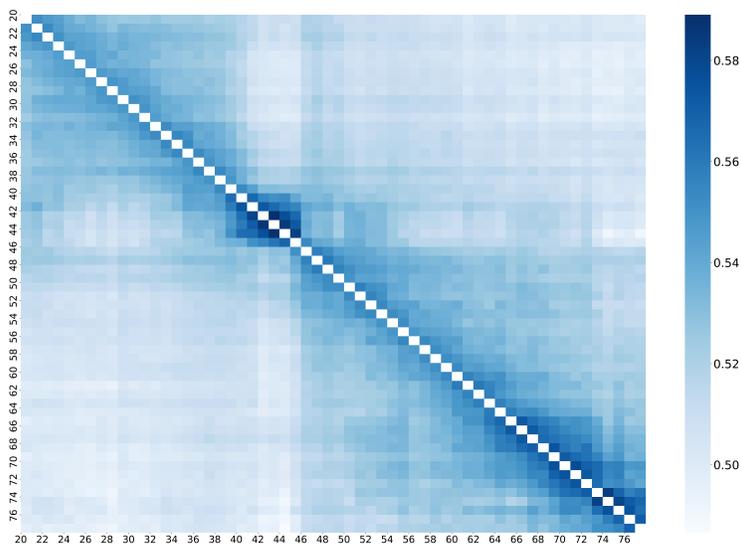}
    \caption{Average similarity between different years of \texttt{HEADLINES}.}
    \label{fig:years}
\end{figure}

Specifically, we first take a random sample of 10,000 texts per year. For year $j$, we embed texts $t_{1j}...t_{10,000j}$ using all-mpnet-base-v2. We choose MPNet because it has been shown to perform well across a variety of embedding datasets and tasks \cite{muennighoff2022mteb}. For each of these $t_{ij}$, we compute the most similar embedding in year $k$, measured by cosine similarity. This gives us a vector of similarity measures $s_{1jk}...s_{10,000jk}$, that for each text in year $j$ measure proximity to the most similar text in year $k$. We average these similarities to calculate $SIM_{jk}$.\footnote{We end in 1977, as we have very few texts after this date.}
Figure \ref{fig:years}, which plots the $SIM_{jk}$, shows that similarity increases with temporal proximity. The dark square towards the upper left is World War 2, during which newspapers coverage was more homogeneous due to the centrality of the war. 

\dataset is useful for training and evaluating models that aim to capture abstractive similarity, whether using the embeddings for tasks like clustering, nearest neighbor retrieval, or semantic search \cite{muennighoff2022mteb}. Because it contains chronological content over a long span of time, it can be used to evaluate dynamic language models for processing continuously evolving content \cite{amba2021dynamic, loureiro2022timelms}, as well as how large language models can be adapted to process historical content \cite{manjavacas2022adapting, ehrmann2021named, manjavacas2021macberth}. Likewise, it can be used to train or evaluate models that predict the region or year a text was written \cite{rastasexplainable}. 
In addition, it is useful for training models and developing benchmarks for a number of downstream tasks, such as topic classification of vast historical and archival documents, which have traditionally been classified by hand. This is an extremely labor-intensive process, and as a result many historical archives and news collections remain largely unclassified. Similarly, it could facilitate creating a large scale dataset to measure term-level semantic change, complementing existing smaller-scale SemEval tasks. 

\dataset has a Creative Commons CC-BY license, to encourage widespread use, and is available on Huggingface.\footnote{https://huggingface.co/datasets/dell-research-harvard/headlines-semantic-similarity}

\section{Existing Semantic Similarity Datasets} \label{sec:lit}

There is a dense literature on semantic similarity, with datasets covering diverse types of textual similarity and varying greatly in size. The focus of \dataset on semantic similarity in historical texts sets it apart from other widely used datasets. It also dwarfs the size of most existing datasets, aggregating the collective work of 20th century newspapers editors, from towns across the U.S. Its paired headlines summarize the same text, rather than being related by other forms of similarity frequently captured by datasets, such as being in the same conversation thread or answering a corresponding question.  

One related class of semantic similarity datasets consists of duplicate questions from web platforms, \textit{e.g.}, questions tagged by users as duplicates from WikiAnswers (77.4 million positive pairs) \cite{fader2014open}, duplicate stack exchange questions (around 304,000 duplicate title pairs) \cite{stackexchange}, and duplicate quora questions (around 400,000 duplicate pairs) \cite{quora}.\footnote{Dataset sizes are drawn, when applicable, from a table documenting the training of Sentence BERT \cite{reimers2019sentence}.}
Alternatively, MS COCO \cite{chen2015microsoft} used Amazon's Mechanical Turk to collect five captions for each image in the dataset, resulting in around 828,000 positive caption pairs. In Flickr \cite{young2014image}, 317,695 positive semantic similarity pairs describe around 32,000 underlying images. Like \dataset, positive pairs in these datasets refer to the same underlying content, but are describing an image rather than providing an abstractive summary of a longer text. 
In future work, \dataset could be expanded to include caption pairs describing the same underlying photo wire image, as local papers frequently wrote their own captions. 

Online comment threads have also been used to train semantic similarity models. For example, the massive scale Reddit Comments \cite{henderson2019repository} draws positive semantic similarity pairs from Reddit conversation threads between 2016 and 2018, providing 726.5 million positive pairs. Semantic similarity between comments in an online thread reflects conversational similarity, to the extent the thread stays on topic, rather than abstractive similarity. 
Likewise, question-answer and natural-language inference datasets are widely used for semantic similarity training.
While other datasets exploit abstractive summaries - \textit{e.g.}, Semantic Scholar (S2ORC) has been used to create semantic similarity pairs of the titles and abstracts of papers that cite each other - to our knowledge there are not large-scale datasets with abstractive summaries of the same underlying texts. 

\begin{figure}[ht] 
    \centering
    \includegraphics[width=\linewidth]{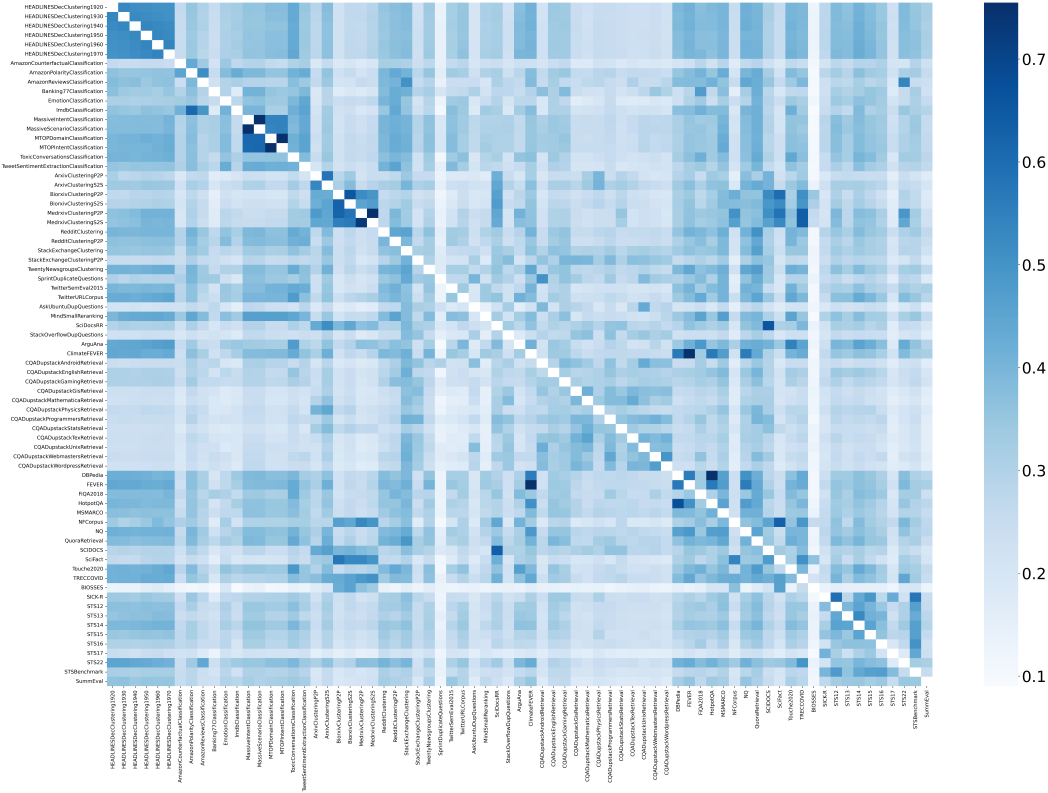}
    \caption{Average similarity between \texttt{HEADLINES} (by decade) and the English datasets in the Massive Text Embedding Benchmark. The similarity measure is described in the text.}
    \label{fig:compare}
\end{figure}

A wide variety of text embedding datasets have been combined into the Massive Text Embedding Benchmark (MTEB) \cite{muennighoff2022mteb}, which evaluates 8 embedding tasks on 58 datasets covering 112 languages. We  measure the similarity between \dataset and the English datasets in MTEB, using the Earth Mover-style distance, described in Section \ref{sec:dataset}. As above, we first take a random sample of (up to) 10,000 texts from each decade of \dataset, as well as each of the English datasets in MTEB (if the dataset contains fewer than 10K texts, we use the full dataset and limit the comparison dataset to the same number of randomly selected texts). For dataset $j$, we embed texts $t_{1j}...t_{10,000j}$. For each of these $t_{ij}$, we compute the most similar embedding in dataset $k$, averaging these across all texts in $j$ to compute $SIM_{jk}$.
$SIM_{jk}$ need not be symmetric. Suppose dataset $j$ is highly homogeneous, whereas dataset $k$ is heterogeneous. $SIM_{jk}$ may be high, because the similar embeddings in homogeneous dataset $j$ are close to a subset of embeddings in dataset $k$. On the other hand, $SIM_{kj}$ may be low, because most texts in dataset $k$ are dissimilar from texts in homogeneous dataset $j$. 

Figure \ref{fig:compare} shows the entire similarity matrix between \dataset and the English datasets in MTEB; rows are the query dataset and columns are the key. The style of the figure was adapted from \cite{muennighoff2022mteb}.  The average similarity between \dataset and MTEB datasets is 31.4, whereas the average similarity between MTEB datasets and \dataset is 33.5, supporting our supposition that \dataset is diverse on average relative to other benchmarks. This average max similarity shows that there is ample information in \dataset not contained in existing datasets. 

Table \ref{ExampleSim} shows examples where the nearest text to a headline in an MTEB dataset is highly similar, versus of average similarity. In some cases, highly similar texts in fact have a different meaning, but the limited context in the MTEB datasets makes this difficult to capture. 

\begin{table}[ht]
    \centering
    \resizebox{.9\linewidth}{!}{
    \begin{threeparttable}
       \begin{tabular}{llc}
      \toprule
\textbf{Headline} & \textbf{Highly similar texts} & \textbf{Similarity} \\ \\
 “Inflation Cuts are Questioned” & \textbf{Reddit}: “Today FOMC Resumed Meeting Inflation & 0.55 \\
 &  getting out of hand… Maybe… Maybe Not” \\
 “Bear Bites Off Arm of Child” & \textbf{StackExchange}: “How to handle animal AI biting & 0.46 \\
 & and holding onto the character... \\
 “British Cruiser Reported Sunk” & \textbf{Twitter}: “That’s It, Britain is Sunk”  & 0.61 \\
 “Will Free Press Dance to Government Tune”  & \textbf{Twitter}: “Donald Trump v. a free press” & 0.60 \\ 
 “Partitioning Plan Unsatisfactory, Ike Declares” & \textbf{Ubuntu Questions}: “Partitioning Issues” & 0.51 \\ 
\toprule
\textbf{Headline} &  \textbf{Average similarity texts} & \textbf{Similarity}\\ \\
 “Reds Knot Strong Tie” & \textbf{ArXiv}: “Knots and Polytopes” & 0.34 \\ 
 “SHOWERS PROMISED TO END HEAT WAVE  & \textbf{StackOverflow}: “how to annotate heatmap with  & 0.27 \\ Two Deaths From Heat Over Weekend”  & text in matplotlib?”  \\ 
“Salary Boost Due For Some On Labor Day”   & \textbf{Twitter}: “Glassdoor will now tell you if youre  & 0.31  \\
& being underpaid” \\ 
“40 Old Ladies Now In Senate Says Rogers”  & \textbf{Quora}: “Is 19 young?” & 0.28 \\




    \bottomrule 
    \end{tabular}
    \end{threeparttable}
  }
    \caption{This table shows similarities between example texts.}
      \label{ExampleSim}
\end{table}

\section{Dataset Construction and Evaluation} \label{sec:methods}

\subsection{Digitization}
We digitized front pages from off-copyright newspapers spanning 1920-1989. 
We recognize layouts using Mask RCNN \cite{he2017mask} and OCR the texts.
We transcribed the headlines using Tesseract. The digitization was performed using Azure F-Series CPU nodes. 

We evaluate this OCR on a hand annotated sample of 300 headlines per decade. Table \ref{tab_desc} reports the character error rate, defined as the Levenshtein distance between the transcribed text and the ground truth, normalized by the length of the ground truth. As expected, OCR quality improves over time, as there is less damage from aging and fewer unusual fonts.

\subsection{Article Association}
\begin{figure}[ht]
     \centering
     \begin{subfigure}[b]{0.2\textwidth}
         \centering
         \includegraphics[width=\textwidth]{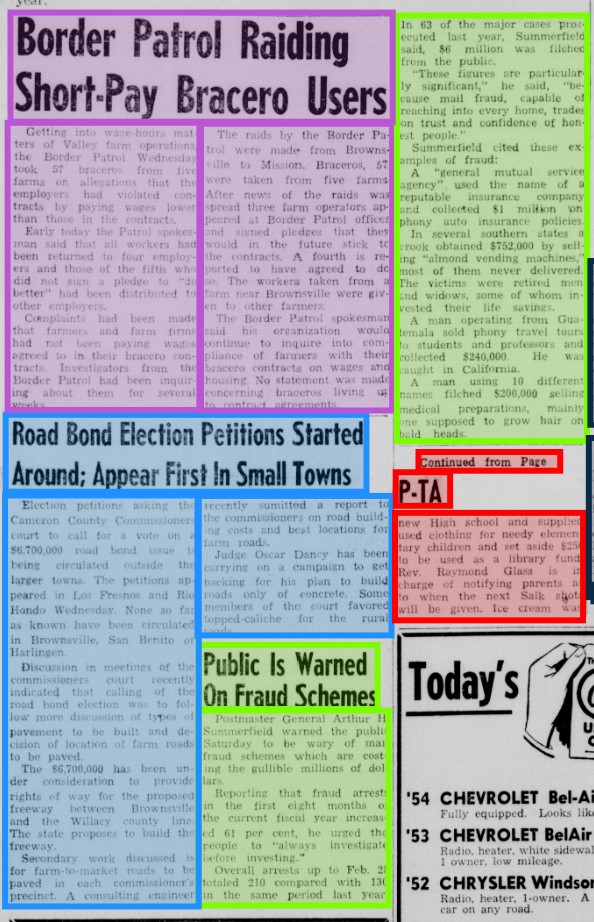}
         \caption{\label{fig:discont}The green article is not-contiguous and it is unclear if the second part continues the purple, green, or even blue articles if the text is not used.}
     \end{subfigure}
     \hfill
     \begin{subfigure}[b]{0.2\textwidth}
         \centering
         \includegraphics[width=\textwidth]{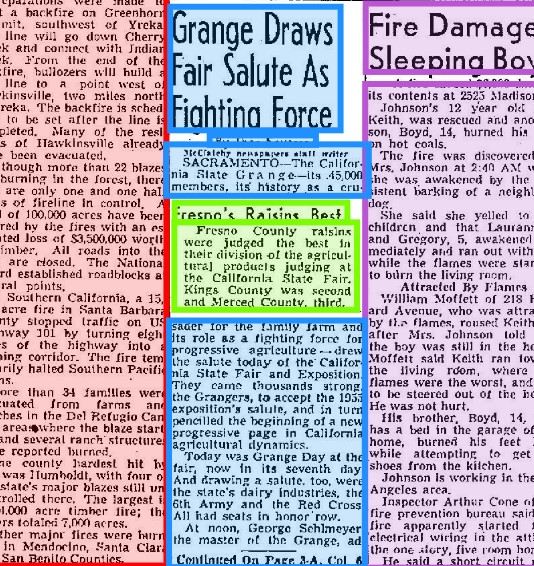}
         \caption{\label{fig:nested}The green article is nested inside the blue article, which causes rule-based and image-based models to either associate the green and blue articles together, or miss the second half of the blue article.}
     \end{subfigure}
          \hfill
     \begin{subfigure}[b]{0.2\textwidth}
         \centering
         \includegraphics[width=\textwidth]{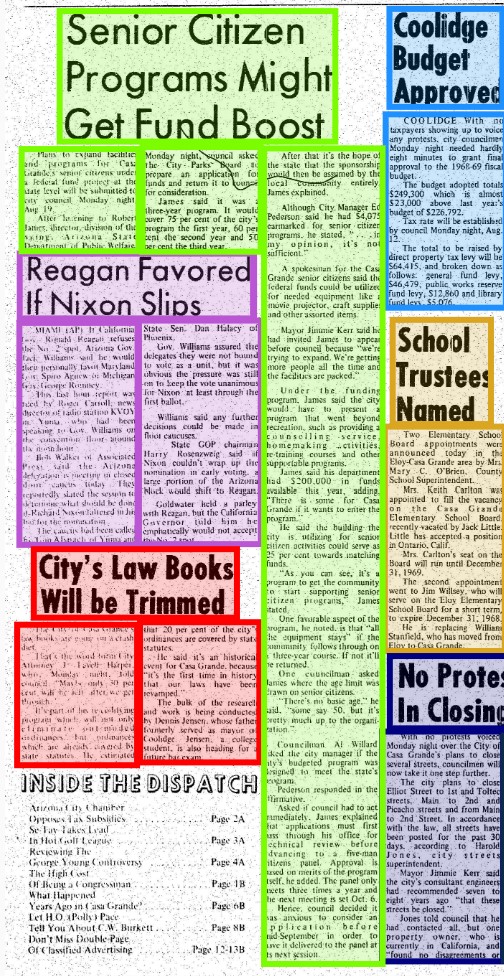}
         \caption{\label{fig:not_sqaure}The smallest rectangle that contains the green article also contains the purple and red articles.}
     \end{subfigure}
    \caption{\label{fig:Articleassocaition} Articles that are misassociated with rule-based or image-based methods}
\end{figure}

Newspaper articles have complex and irregular layouts that can span multiple columns (Figure \ref{fig:Articleassocaition}). 

We associate the (potentially multiple) headline bounding boxes with the (potentially multiple) article bounding boxes and byline boxes that comprise a single article using a combination of layout information and language understanding. 
A rule-based approach using the document layouts gets many of the associations correct, but misses some difficult cases where article bounding boxes are arranged in complex layouts. Language understanding can be used to associate such articles but must be robust to noise, from errors in layout detection (\textit{e.g.} from cropping part of a content bounding box or adding part of the line below) and from OCR character recognition errors. 

\begin{figure}[ht]
     \centering
     \begin{subfigure}[b]{0.3\textwidth}
         \centering
         \includegraphics[width=\textwidth]{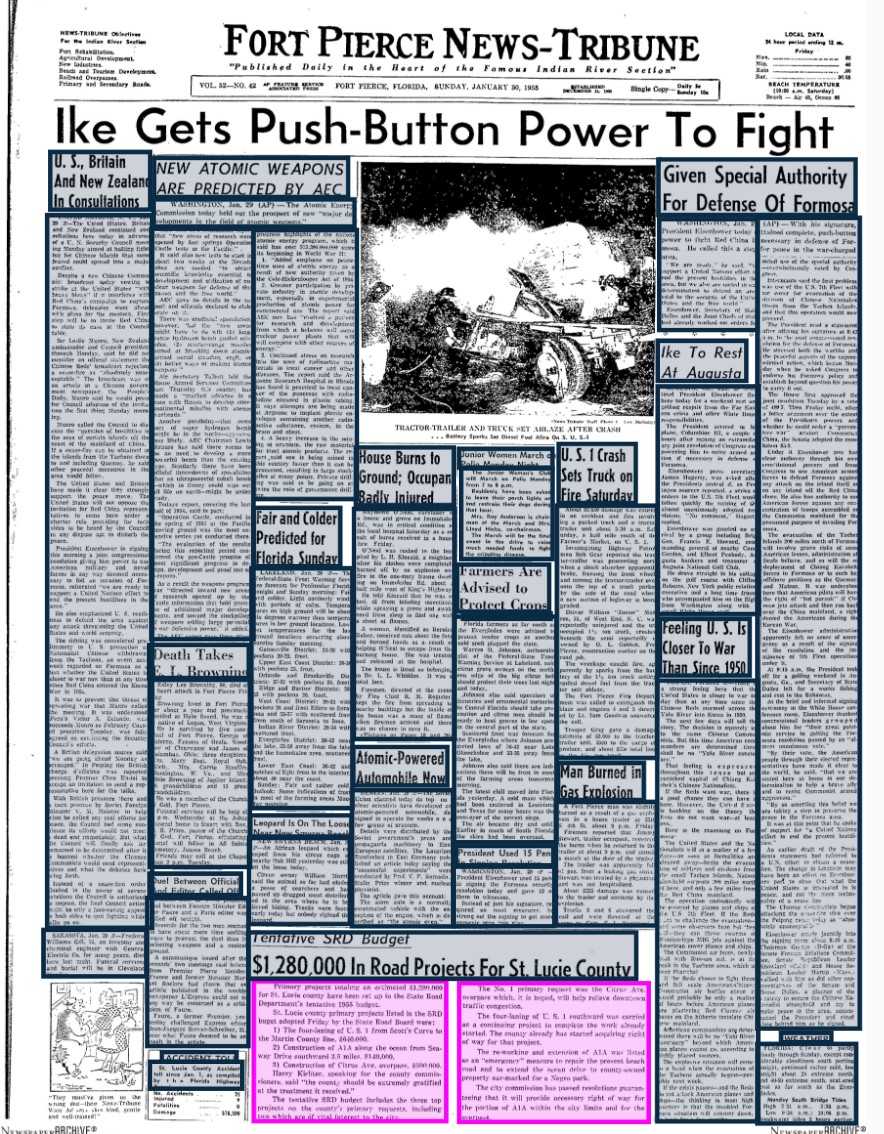}
         \caption{Article bounding boxes that are under the same headline and at the bottom of the page are used to create training data.}
     \end{subfigure}
     \hfill
     \begin{subfigure}[b]{0.3\textwidth}
         \centering
         \includegraphics[width=\textwidth]{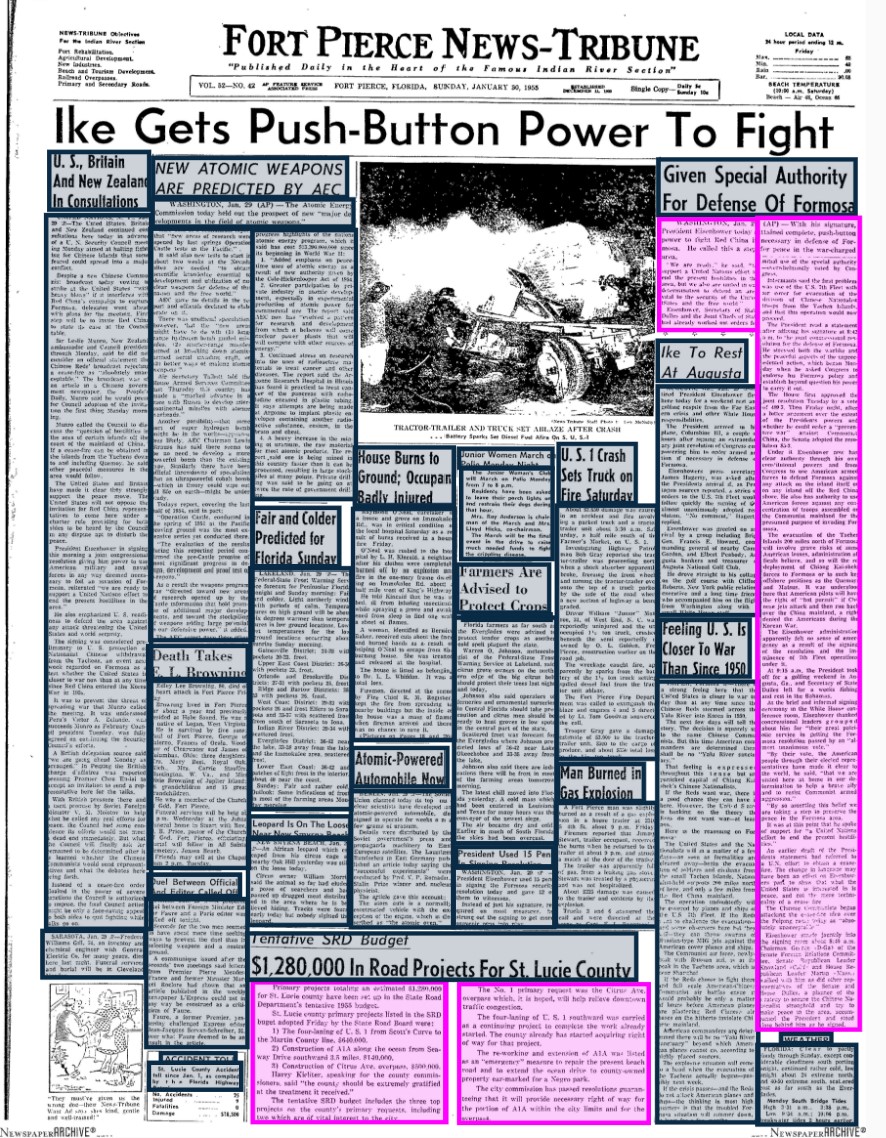}
         \caption{At inference time, all bounding boxes that are directly under the same headline are associated with each other.}
     \end{subfigure}
          \hfill
     \begin{subfigure}[b]{0.3\textwidth}
         \centering
         \includegraphics[width=\textwidth]{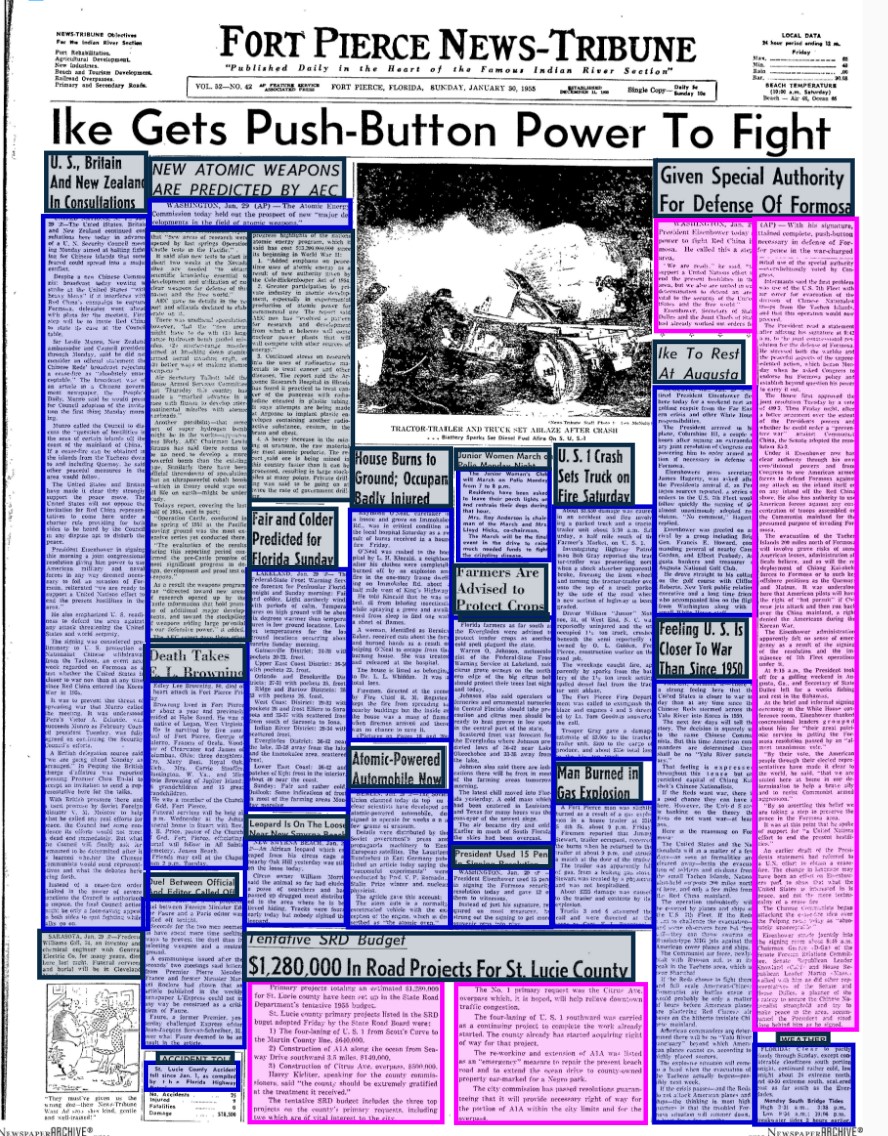}
         \caption{Any other article with a headline directly above it is not compared, leaving only a few orphans that are left to be associated.}
     \end{subfigure}
     \begin{subfigure}[b]{0.3\textwidth}
         \centering
         \includegraphics[width=\textwidth]{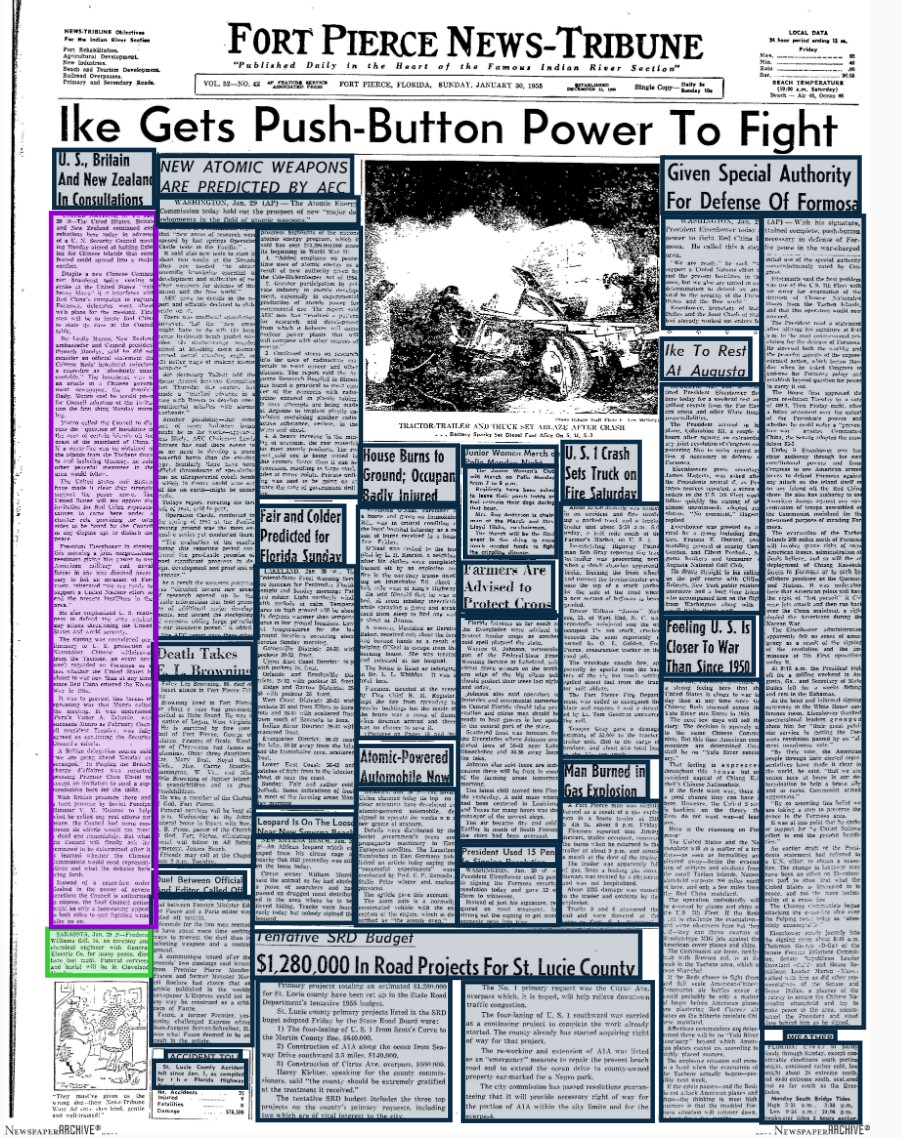}
         \caption{This orphan (green bounding box) is compared with the bounding box above it. In this case, it is a separate article without a headline so it is not associated.}
     \end{subfigure}
     \hfill
     \begin{subfigure}[b]{0.3\textwidth}
         \centering
         \includegraphics[width=\textwidth]{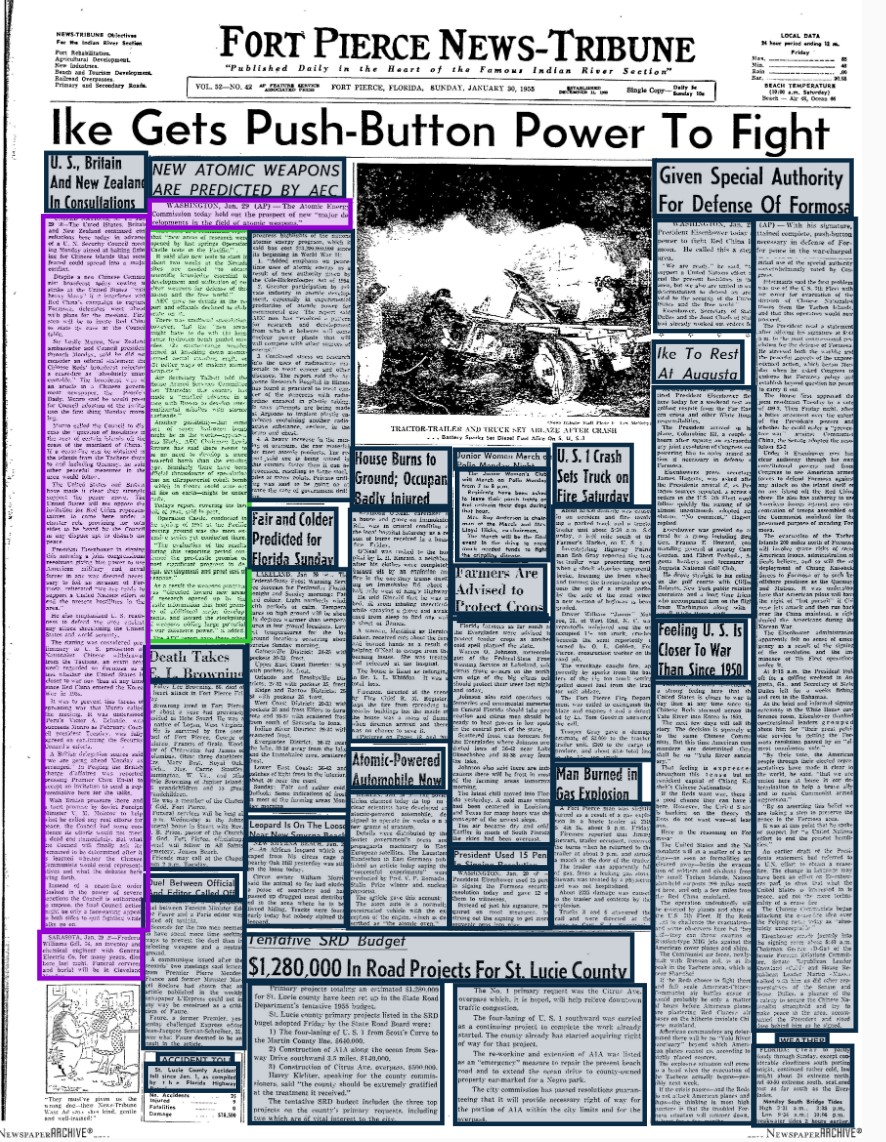}
         \caption{Orphans are also compared with bounding boxes to the right. This orphan (green bounding box) is associated with the bounding box directly above.}
     \end{subfigure}
          \hfill
     \begin{subfigure}[b]{0.3\textwidth}
         \centering
         \includegraphics[width=\textwidth]{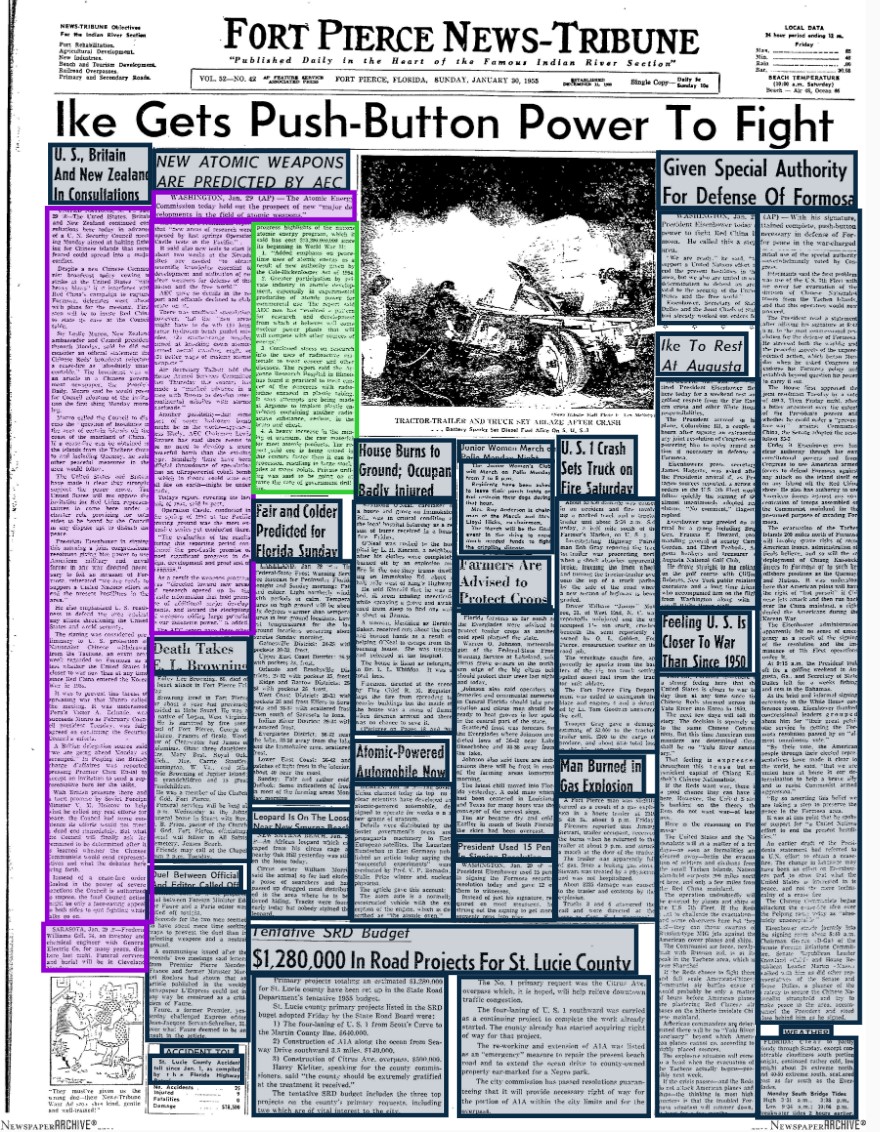}
         \caption{The final orphan is compared with two columns to the right, as sometimes articles skip columns. If further columns to the right exist, they are not compared.}
     \end{subfigure}
    \caption{Illustration of article association pipeline}
    \label{rules}
\end{figure}

Hand annotating a sufficiently large training dataset would have been infeasibly costly. Instead, we devise a set of rules that - while recall is relatively low - have precision above 0.99, as measured on an evaluation set of 3,803 labeled bounding boxes. The algorithm exploits the positioning of article bounding boxes relative to headline boxes (as in Figure \ref{rules}), first grouping an article bounding box with a headline bounding box if the rules are met and then associating all article bounding boxes grouped with the same headline together. Since precision is above 0.99, the rule generates nearly perfect silver-quality training data.

To train a language model to predict whether article box B follows box A, we embed the first and last 64 tokens of the texts in boxes B and A, respectively, with a RoBERTa base model \cite{liu2019roberta}. The pair is positive when B follows A. The training set includes 12,769 positive associated pairs, with training details described in the supplementary materials. 
At inference time, we first associate texts using the rule-based approach, described in  Figure \ref{rules}, which has extremely high precision. To improve recall, we then apply the RoBERTa cross-encoder to remaining article boxes that could plausibly be associated, given their coordinates. Texts cannot be followed by a text that appears to the left, as layouts always proceed from left to right, so these combinations are not considered. 

We  evaluate this method on a hand-labeled dataset of 214 scans. Full details of this dataset are given in the appendix. Table \ref{pipelineEval} evaluates recall, precision and F1 for associated articles. The F1 of 93.7 is high, and precision is extremely high. Errors typically occur when there is an error in the layout analysis or when contents are very similar, \textit{e.g.}, grouping multiple obituaries into a single article.

\begin{table}[ht]
    \centering
    \resizebox{.55\linewidth}{!}{
    \begin{threeparttable}
       \begin{tabular}{lccc}
      \toprule
 & (1) & (2) & (3)  \\
	&	F1 & Recall & Precision \\
 \cmidrule{2-4}
Full Article Association & 93.7 & 88.3 & 99.7 \\
     \bottomrule 
    \end{tabular}
    \end{threeparttable}
  }
    \caption{This table evaluates the full article association model.}
      \label{pipelineEval}
\end{table}

\subsection{Detecting Reproduced Content}

Accurately detecting reproduced content can be challenging, as articles were often heavily abridged by local papers to fit within their space constraints and errors in OCR or article association can add significant noise. Table \ref{WireClusters} shows examples of reproduced articles.

\begin{table}[ht]
\begin{small}
    \centering
    {\renewcommand{\arraystretch}{2}%
    \begin{tabular}{p{0.15\linewidth} p{0.15\linewidth} p{0.15\linewidth}  p{0.15\linewidth} p{0.15\linewidth}  p{0.15\linewidth}}
    
    \raisebox{-\totalheight}{\includegraphics[width=\linewidth]{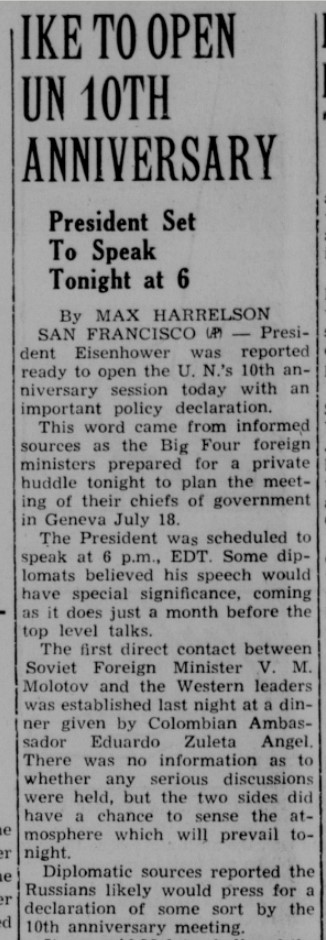}} & 
    
    \tiny{SAN FRANCISCO (\underline{\#} — Presi\underline{-}dent Eisenhower was\underline{\_} reported ready to open the U. N.’s 10th anniversary session today with an important policy declaration. 
    
    This word came from informed sources as the Big Four foreign ministers prepared for a private huddle tonight to plan the meeting of their chiefs of government in Geneva July 18. 
    
    \hl{The President was scheduled to speak at 6 p.m., EDT.} Some diplomats believed his speech would have special significance, coming as it does just a month before the top level talks. 
    
    The first direct contact between Soviet Foreign Minister V. M. Molotov and the Western leaders was established last night at a dinner given by Colombian Ambassador Eduardo Zuleta Angel \hl{There was no information as tc whether any serious discussion: were held, but the two sides dic have a chance to sense the at-'\}mosphere which will prevail to-'| night..|}
    
    Diplomatic sources reported \underline{th«} Russians likely would press for \underline{: 'Ideclaration} of some sort by the \underline{|} 10th anniversary meeting}
    
    &  \raisebox{-\totalheight}{\includegraphics[width=\linewidth]{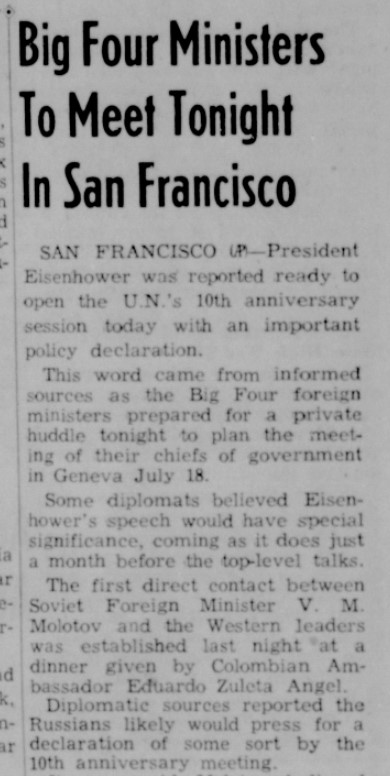}} & 

    \tiny{SAN FRANCISCO \underline{\#}—President Eisenhower was reported ready to open the \underline{U\_N.}'s 10th anniversary session today with an important policy declaration.
    
    This word came from informed sources as the Big Four foreign ministers prepared for a private huddle tonight to plan the meeting of their chiefs of government in Geneva July 18.
    
    Some diplomats believed Eisenhower’s speech would have special significance, coming as it does just \underline{'}a month before the top-level talks.\underline{|} The first direct contact between (Soviet Foreign Minister V. M.\underline{|} Molotov and the Western leaders \underline{|} was established last night “at a\underline{'}dinner given by Colombian \underline{Am| bassador} Eduardo \underline{Zulcta} Angel.
    
    \underline{|} Diplomatic sources reported the \underline{'}Russians likely would press for a declaration of some sort by the \underline{110th} anniversary meeting.}
     
     & \raisebox{-\totalheight}{\includegraphics[width=\linewidth]{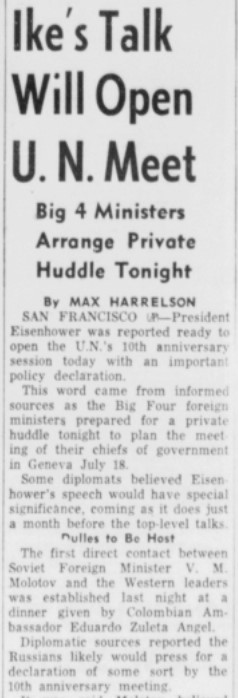}} & 

     \tiny{"\underline{SY MAA} \underline{MARKRKRELSUN} SAN FRANCISCO \underline{\#®}—President Eisenhower was reported ready to open the U.N.’s 10th anniversary session today with an \underline{importan!} policy declaration\underline{,}
     
     This word came from informed sources as the Big Four foreign ministers prepared for a private huddle tonight to plan the meeting of their chiefs of government in Geneva July 18.
     
     Some diplomats believed \underline{Eisen.hower’s} speech would have special\underline{!} significance, coming as it does just a month before the top-level talks.
     
     \hl{Mulles to Bo Host}
     
     The first direct contact between Soviet Foreign Minister V. M. Molotov and the Western leaders was established last night at a dinner given by Colombian Ambassador Eduardo Zuleta Angel.
     
     Diplomatic sources reported the Russians likely would press for a \underline{'}declaration of some sort by the 10th anniversary meeting.}\\

    \end{tabular}}
    \caption{Examples of reproduced articles. Additions are highlighted, and OCR errors are underlined.}
    \label{WireClusters}
\end{small}
\end{table}

We use the model developed by \cite{silcock2023noise}, who show that a contrastively trained neural MPNet bi-encoder - combined with single linkage clustering of article representations - accurately and cheaply detects reproduced content. 
This bi-encoder is contrastively trained on a hand-labeled dataset (detailed in the appendix) to create similar representations of articles from the same wire source and dissimilar representations of articles from different underlying sources, using S-BERT's online contrastive loss \cite{hadsell2006dimensionality} implementation. 

We run clustering on the article embeddings by year over all years in our sample. 
In post-processing, we use a simple set of rules exploiting the dates of articles within clusters to remove content like weather forecasts and legal notices, that are highly formulaic and sometimes cluster together when they contain very similar content (\textit{e.g.} a similar 5-day forecast) but did not actually come from the same underlying source. We remove all headline pairs that are below a Levenshtein edit distance, normalized by the min length in the pair, of 0.1, to remove pairs that are exact duplicates up to OCR noise. 
Training and inference were performed on an A6000 GPU card. 
More details are provided in the supplementary materials. 

To evaluate how well the model detects reproduced content, we use a labeled sample of all front page articles appearing in the newspaper corpus for three days in the 1930s and 1970s, taken from \cite{silcock2023noise}. This sample consists of 54,996 positive reproduced article pairs and 100,914,159 negative pairs. The large-scale labeled evaluation dataset was generated using the above pipeline, so the evaluation is inclusive of any errors that result from upstream layout detection, OCR, or article association errors. 

The neural bi-encoder methods achieve a high adjusted rand index (ARI) of 91.5, compared to 73.7 for an optimal local sensitive hashing specification, chosen on the validation set. This shows that our neural methods substantially outperform commonly used sparse methods for detecting reproduced content. The neural bi-encoder is slightly outperformed by adding a re-ranking step that uses a neural cross-encoder on the best bi-encoder matches (ARI of 93.7). We do not implement this method because the cross-encoder doesn't scale well. In contrast, the bi-encoder pipeline can be scaled to 10 million articles on a single GPU in a matter of hours, using a FAISS \cite{johnson2019billion} backend. 

\begin{table}[ht]
    \centering
    \begin{threeparttable}
        \begin{tabular}{l|c|c}
        \toprule
             &  \textbf{Neural} & \textbf{Non-Neural} \\
             \midrule
            \textbf{Most scalable} & Bi-encoder (91.5) & LSH (73.7)  \\
            \textbf{Less scalable} & Re-ranking (\textbf{93.7}) & $N$-gram overlap (75.0) \\
        \bottomrule 
    \end{tabular}
    \end{threeparttable}
    \caption{The numbers in parentheses are the Adjusted Rand Index for four different models - a bi-encoder, a ``re-ranking'' strategy that combines a bi- and cross-encoder, locally sensitive hashing (LSH), and $N$-gram overlap. Hyperparameters were chosen on the \texttt{NEWS-COPY} validation set, and all models were evaluated on the \texttt{NEWS-COPY} test set.}
    \label{table:wire_eval}
    \vspace{-5mm}
\end{table}

An error analysis is provided in \cite{silcock2023noise}. 
Errors typically consist of articles about the same story from different wire services (\textit{e.g.} the Associated Press and the United Press) or updates to a story as new events unfolded. Both types of errors will plausibly still lead to informative semantic similarity pairs.

 
\section{Benchmarking} \label{sec:benchmark}

We benchmark \dataset using a variety of different language models and the MTEB clustering task. This task embeds texts using different base language models and then uses $k$ - the number of clusters in the ground truth data - for k-means clustering. Following MTEB, we score the model using the v-measure \cite{rosenberg2007v}. We should note that real-world problems are often framed as clustering tasks - rather than as classification tasks - because $k$ is unknown. By using $k$ from the ground truth, it makes the task easier. Nevertheless, we examine this task to allow for comparison with the rest of the literature. 

\begin{figure}[ht] 
    \centering
    \includegraphics[width=.7\linewidth]{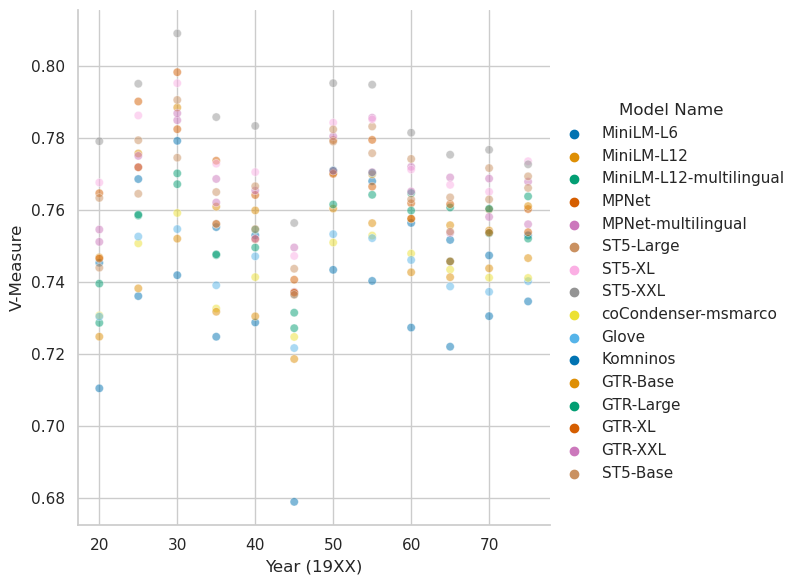}
    \caption{This figure benchmarks \texttt{HEADLINES} on the MTEB clustering task. The x-axis shows the year that the sample was taken from and the y-axis gives the v-measure.}
    \label{fig:benchmark}
\end{figure}

Figure \ref{fig:benchmark} plots the results of this benchmarking exercise. 
MTEB benchmarks clustering on Arxiv, Bioarxiv, Medarxiv, Reddit, StackExchange, and Twenty Newsgroups. Texts are labeled with their classification (e.g., fields like ComputerVision for the Arxiv datasets; the subreddit for Reddit). The best average v-score across these datasets, from MPNet, is 43.69. The best average v-score across decades for \dataset, from ST5-XXL, is around 78. This difference is likely to reflect, at least in part, that our cluster labels are less noisy, since texts in the same cluster summarize the same content. In contrast, titles of Reddit posts in the same subreddit may be only loosely linked to each other, and many could be within the domain of another subreddit cluster. While a user happened to post in one subreddit, another user could have reasonably made the same titled post in a different subreddit. The under-identification of the clustering tasks for some texts in the MTEB datasets is suggested by the very low v-scores across state-of-the-art language models. Overall, this suggests the high quality of clusters in \dataset relative to many web text datasets. Yet there is still ample scope for improvement to the state-of-the-art model.

\section{Limitations and Recommended Usage} \label{sec:limits}

\dataset contains some transcription errors. For working with historical texts, these are more a feature than a bug, as most historical texts are transcribed and also contain various OCR errors. Training a model on transcribed texts likely makes it more robust to transcription errors at inference time. However, researchers requiring completely clean texts should seek another corpus. 

\dataset contains historical language, that reflects the semantics and cultural biases of many thousands of local newspaper editors. This is a distinguishing feature of \dataset, that is core to many potential applications. We do not attempt to filter texts with antiquated terms or that may be considered offensive, as this would invalidate the use of the dataset for studying semantic change and historical contexts. 
At the same time, this makes \dataset less suited for tasks that require texts that fully conform to current cultural standards or semantic norms. For these reasons, we recommend against the use of \dataset for training generative models. 
Rather, with nearly 400M positive semantic similarity pairs spanning much of the 20th century, it can plausibly play an important role in facilitating the application of large language models to historical texts.

\clearpage

\setcounter{table}{0}
\renewcommand{\thetable}{S-\arabic{table}} 
\setcounter{figure}{0}
\renewcommand{\thefigure}{S-\arabic{figure}} 
\setcounter{section}{0}
\renewcommand{\thesection}{S-\arabic{section}}

\begin{center}
    \section*{Supplementary Materials}

\end{center}

\section{Methods to Associate Articles}

Figure 6 in the main text illustrates the full article association procedure. 

First, we used a rule-based algorithm using associate article bounding boxes that are under the same headline, as these are part of the same article with extremely high probability. Algorithm \ref{rule_algo} gives pseudocode for this method. We set the parameters as $P_S = 100$, $P_T = 20$, $P_B = 50$.

For training data, where we want article pairs that are not only part of the same article, but also where they appear in the given order, we further narrow down the pairs. Specifically, we use only those pairs which are horizontally next to each other, and which have no other bounding boxes below them, as for these pairs, we can guarantee that the pair of bounding follow directly after one another (whereas for other article bounding boxes that share a headline, there may be a third bounding box in between). Algorithm \ref{order_algo} shows pseudocode for this procedure, and we used $P_C = 5$, and it is further illustrated in panel A of figure 6 in the main text. 

For hard negatives, we used article boxes under the same headline in reverse reading order (right to left). For standard negatives, we took pairs of articles on the same page, where B was above and to the left of A, as articles do not read from right to left. One twelfth of our training data were positive pairs, another twelfth were hard negative pairs and the remainder were standard negative pairs. This outperformed a more balanced training sample. 

We use this dataset to finetune a cross-encoder using a RoBERTa base model \cite{liu2019roberta}. We used a Bayesian search algorithm \cite{pmlr-v80-falkner18a} to find optimal hyperparameters on one tenth of our training data (limited compute prevented us from running this search with the full dataset), which led to a learning rate of 1.7e-5, with a batch size of 64 and 29.57\% warm up. We trained for 26 epochs with an AdamW optimizer, and optimize a binary cross-entropy loss.

We evaluate these methods on a hand-labeled dataset of 214 scans, randomly selected from 1968 and 1955. These scans were labeled by a highly-trained undergraduate research assistant. Summary statistics of this dataset are given in table \ref{tab_aa_dat} and evaluation results are given in the main text. 
\begin{table}[ht]
    \centering
    \resizebox{.8\linewidth}{!}{
    \begin{threeparttable}
\begin{tabular}{@{}cccc@{}}
\toprule
Scan count & Article bounding boxes & Headline bounding boxes & Article-article associations \\ \midrule
214 & 3,803 & 2,805 & 1,851 \\ \bottomrule
\end{tabular}
    \end{threeparttable}
  }
    \caption{Descriptive statistics of article association training data.}
      \label{tab_aa_dat}
\end{table}

\begin{algorithm}[ht!]
\caption{Rule-based association of article bounding boxes}\label{rule_algo}

\begin{algorithmic}[1]
\INPUT $b_1, ..., b_n \in B$: set of bounding boxes that appear on the same scan, with their
coordinates, denoted $left(b_i)$, $right(b_i)$, $top(b_i)$, $bottom(b_i)$, 
and type (headline, article, byline etc.), denoted $type(b_i)$.

\PARAMETERS 
\Statex $W$: width of scan
\Statex $H$: height of scan
\Statex $P_S$: fraction of width, for creating side margin
\Statex $P_T$: fraction of height, for creating top margin
\Statex $P_B$: fraction of height, for creating bottom margin 

\OUTPUT $ArticleArticlePairs = \{(b_i, b_j) \in B \times B | b_i, b_j$ predicted to be part of the same full article and $type(b_i) = type(b_j) = article\}$ 

\State Initialise: $M_S = W/P_S$, the side margin, $M_B = H/P_B$, the bottom margin, $M_T = H/P_T$, the top margin, $MatchedHeadlines = \{\}$, $HeadlineArticlePairs = \{\}$, $ArticleArticlePairs = \{\}$

\For{all $b_0$ in B where $type(b_0)$ is article}

\State 	Create $B_0 \subset B$ where:

    \Statex \hspace{2em} - All bounding boxes of type byline are removed 
    \Statex \hspace{2em} - $b_0$ is removed
    \Statex \hspace{2em} - All bounding boxes are removed that  do not share at least $M_{S}$ of the horizontal axis     
    \Statex \hspace{2em} - All bounding boxes are removed whose bottom is more than $M_{B}$ below the top of $b_0$ 
    \Statex \hspace{2em} - All bounding boxes are removed whose bottom is more than $M_{T}$ above the top of $b_0$ 

\If{$B_0$ is not empty} 

\State Let $b_1$ be the element of $B_0$ that has the lowest bottom coordinate 

\If{$type(b_1)$ is headline} 

\State $MatchedHeadlines = MatchedHeadlines \cup \{b_1\}$
\State $HeadlineArticlePairs = HeadlineArticlePairs \cup \{(b_0, b_1)\}$
\EndIf
\EndIf
\EndFor

\For{all $b_h$ in $MatchedHeadlines$}

\State Let $H_1 \subset HeadlineArticlePairs$ be all pairs that contain that headline, $b_h$

\If {$H_1$ has at least two elements}

\State Let $A$ be all the bounding boxes of type article from the pairs in $H_1$
\State Let $C$ be all combinations of 2 elements of $A$
\State $ArticleArticlePairs = ArticleArticlePairs \cup  C$

\EndIf
\EndFor

\end{algorithmic}
\end{algorithm}

\begin{algorithm}[ht!]
\caption{Selection of ordered article pairs }\label{order_algo}

\begin{algorithmic}[1]
\INPUT  $ArticleArticlePairs$, $B$, from algorithm \ref{rule_algo}.

\PARAMETERS 
\Statex $P_C$: fraction of column width, for creating margin

\OUTPUT $OrderedPairs \subset ArticleArticlePairs$ 

\State Initialise: $OrderedPairs = \{\}$

\For{$p$ in $ArticleArticlePairs$}
\State 	Let $p_l$ be the element of $p$ with the furthest left coordinate
\State 	Let $p_r$ be the other element 

\If {$left(p_r)$ is not further to the right of $right(p_l)$ than $width(p_l)/F_C$}

\If {there are no other bounding boxes below $p_l$} 
\State			$OrderedPairs = OrderedPairs \cup \{p\}$ 

\EndIf
\EndIf
\EndFor

\end{algorithmic}
\end{algorithm}

\section{Methods to Detect Reproduced Content}

To detect reproduced content, we  use the contrastively trained bi-encoder model developed by \cite{silcock2023noise}, which is trained to learn similar representations for reproduced articles and dissimilar representations for non-reproduced articles. This model is based on an S-BERT MPNET model \cite{reimers2019sentence, song2020mpnet} and is fine-tuned on a hand-labelled dataset of articles from the same underlying wire source, using S-BERT's online contrastive loss \cite{hadsell2006dimensionality} implementation, with a 0.2 margin and cosine similarity as the distance metric.
The learning rate is 2e-5 with 100\% warm up and a batch size of 32. It uses an AdamW optimizer, and the model is trained for 16 epochs. This bi-encoder is trained and evaluated on a hand-labeled dataset, which is detailed in \ref{table:wiresumstat}. The results of this evaluation are given in the main text. 

To create clusters from the bi-encoder embeddings, we use highly scalable single-linkage clustering, with a cosine similarity threshold of 0.94. We build a graph using articles as nodes, and add edges if the cosine similarity is above this threshold. As edge weights we use the negative exponential of the difference in dates (in days) between the two articles. We then apply Leiden community detection to the graph to control false positive edges that can otherwise merge disparate groups of articles.

We further remove clusters that have over 50 articles and contain articles with greater than five different dates. We also remove clusters that contain over 50 articles, when the number of articles is more than double the number of unique newspapers from which these articles are sourced. This removes clusters of content that are correctly clustered in the sense of being based on the same underlying source, but are not useful for the \dataset dataset. For example, an advertisement (misclassified as an article due to an article-like appearance) might be repeated by the same newspaper on multiple different dates and would be removed by these rules, or weather forecasts can be very near duplicates across space and time, forming large clusters. 

\begin{table}[t]
    \centering
    \begin{threeparttable}
        \begin{tabular}{lccccc}
        \toprule
             &  Positives & Negative  & Reproduced  & Singleton & Total    \\
             & Pairs & Pairs & Articles & Articles & Articles \\
        \midrule
        \textbf{Training Data} \\
       Training	&	36,291	&	37,637	&	891	&	--	&	7,728 \\
Validation	&	3,042	&	3,246	&	20	&	--	&	283 \\
\midrule
\textbf{Full Day Evaluation}	&		&		&		&		&	\\
Validation	&	28,547	&	12,409,031	&	447	&	2,162	&	4,988 \\
Test	&	54,996	&	100,914,159	&	1,236	&	8,046	&	14,211 \\
\midrule
\textbf{Full Dataset}	&	122,876	&	113,364,073	&	2,594	&	10,208 	&	27,210 \\
        \bottomrule 
    \end{tabular}
    \end{threeparttable}
    \caption{Summary statistics of training and evaluation data for detecting duplicate content.}
    \label{table:wiresumstat}
    \vspace{-5mm}
\end{table}

\section{A Summary of Copyright Law for Works Published in the United States}


\begin{table*}[h]
    \centering
    \resizebox{\linewidth}{!}{
    \begin{threeparttable}
       \begin{tabular}{lcc}
      \toprule
Date of Publication	&	Conditions	&	Copyright Term \\
\cmidrule{1-3}
\textbf{\textit{Public Domain}} \\
Anytime	&	Works prepared by an officer/employee of the	&	None \\ 
 & U.S. Government as part of their official duties \\ \\
\textbf{Before 1928}	&	\textbf{None}	&	\textbf{None. Copyright expired.}	\\ \\
\textbf{1928 through 1977}	&	\textbf{Published without a copyright notice}	&	\textbf{None. Failure to comply with required formalities}	\\ \\
\textbf{1978 to 1 March 1989}	&	\textbf{Published without notice and }	&	\textbf{None. Failure to comply with required formalities}	\\
 & \textbf{without subsequent registration within 5 years} \\ \\
\textbf{1928 through 1963}	&	\textbf{Published with notice}	&	\textbf{None. Copyright expired}	\\
 & \textbf{but copyright was not renewed} \\ \\

\textbf{\textit{Copyrighted}} \\
1978 to 1 March 1989	&	Published without notice, but with 	&	70 (95) years after the death of author (corporate author)	\\
 & subsequent registration within 5 years \\ \\
1928 through 1963	&	Published with notice 	&	95 years after publication	\\
 & and the copyright was renewed \\ \\
1964 through 1977	&	Published with notice	&	95 years after publication 	\\ \\
1978 to 1 March 1989	&	Created after 1977 and published with notice	&	70 (95) years after the death of author (corporate author) \\
 & & or 120 years after creation, if earlier	\\ \\
1978 to 1 March 1989	&	Created before 1978 and first published 	&	The greater of the term specified in the previous entry \\ 
& with notice in the specified period & or 31 December 2047	\\ \\
From 1 March 1989 through 2002	&	Created after 1977	&	70 (95) years after the death of author (corporate author) \\
 & & or 120 years after creation, if earlier	\\ \\
From 1 March 1989 through 2002	&	Created before 1978 and &	The greater of the term specified in the previous entry \\ 
& first published in this period & or 31 December 2047	\\ \\
After 2002	&	None	&	70 (95) years after the death of author (corporate author) \\
& & or 120 years after creation, if earlier \\ \\

     \bottomrule 
    \end{tabular}
    \end{threeparttable}}
    \caption{This table summarizes U.S. copyright law, based on a similar table produced by the Cornell libraries. For concision, we focus on works initially published in the United States. A variety of other cases are also covered at \url{https://guides.library.cornell.edu/copyright}.}
      \label{copyright}
\end{table*}

\newpage
\section{Dataset Information}

\subsection{Dataset URL}

\dataset can be found at \url{https://huggingface.co/datasets/dell-research-harvard/headlines-semantic-similarity}. 

This dataset has structured metadata following \url{schema.org}, and is readily discoverable.\footnote{See \url{https://search.google.com/test/rich-results/result?id=_HKjxIv-LaF_8ElAarsM_g} for full metadata.} 

\subsection{DOI}
The DOI for this dataset is: 10.57967/hf/0751.

\subsection{License}
\dataset has a Creative Commons CC-BY license.

\subsection{Dataset usage}

The dataset is hosted on huggingface, in json format. Each year in the dataset is divided into a distinct file (eg. 1952\_headlines.json).  

The data is presented in the form of clusters, rather than pairs to eliminate duplication of text data and minimize the storage size of the datasets. 

An example from \dataset looks like:  

\begin{verbatim}
{
    "headline": "FRENCH AND BRITISH BATTLESHIPS IN MEXICAN WATERS",
    "group_id": 4
    "date": "May-14-1920",
    "state": "kansas",
}
\end{verbatim}

The data fields are: 

- \verb|headline|: headline text.

- \verb|date|: the date of publication of the newspaper article, as a string in the form mmm-DD-YYYY.

- \verb|state|: state of the newspaper that published the headline.

- \verb|group_id|: a number that is shared with all other headlines for the same article. This number is unique across all year files.

The whole dataset can be easily downloaded using the \verb|datasets| library: 

\begin{verbatim}
from datasets import load_dataset
dataset_dict = load_dataset("dell-research-harvard/headlines-semantic-similarity")
\end{verbatim}

Specific files can be downloaded by specifying them:

\begin{verbatim}
from datasets import load_dataset
load_dataset(
    "dell-research-harvard/headlines-semantic-similarity", 
    data_files=["1929_headlines.json", "1989_headlines.json"]
)
\end{verbatim}

\subsection{Author statement}
We bear all responsibility in case of violation of rights.

\subsection{Maintenance Plan}

We have chosen to host \dataset on huggingface as this ensures long-term access and preservation of the dataset.

\subsection{Dataset documentation and intended uses}
We follow the datasheets for datasets template \cite{gebru2021datasheets}. 

\subsubsection{Motivation}

\paragraph{For what purpose was the dataset created?}{Was there a specific task in mind? Was there a specific gap that needed to be filled? Please provide a description.}

\textit{Transformer language models contrastively trained on large-scale semantic similarity datasets are integral to a variety of applications in natural language processing (NLP). A variety of semantic similarity datasets have been used for this purpose, with positive text pairs related to each other in some way. Many of these datasets are relatively small, and the bulk of the larger datasets are created from recent web texts; e.g. positives are drawn from the texts in an online comment thread or duplicate questions in a forum. Relative to existing datasets, \dataset is very large, covering a vast array of topics. This makes it useful generally speaking for semantic similarity pre-training. It also covers a long period of time, making it a rich training data source for the study of historical texts and semantic change. It captures semantic similarity directly, as the positive pairs summarize the same underlying texts. }

\paragraph{Who created this dataset (e.g., which team, research group) and on behalf of which entity (e.g., company, institution, organization)?} ~\\
\textit{\dataset was created by Melissa Dell and Emily Silcock, at Harvard University.}

\paragraph{Who funded the creation of the dataset?}{If there is an associated grant, please provide the name of the grantor and the grant name and number.}

\textit{The creation of the dataset was funded by the Harvard Data Science Initiative, Harvard Catalyst, and compute credits provided by Microsoft Azure to the Harvard Data Science Initiative.}

\paragraph{Any other comments?} ~\\
\textit{None.
}
\bigskip
\subsubsection{Composition}

\paragraph{What do the instances that comprise the dataset represent (e.g., documents, photos, people, countries)?}{ Are there multiple types of instances (e.g., movies, users, and ratings; people and interactions between them; nodes and edges)? Please provide a description.}

\textit{\dataset comprises instances of newspaper headlines and relationships between them. Specifically, each headline includes information on the text of the headline, the date of publication, and the state it was published in. Headlines have relationships between them if they are semantic similarity pairs, that is, if they two different headlines for the same newspaper article.}

\paragraph{How many instances are there in total (of each type, if appropriate)?}  ~\\
\textit{\dataset contains 34,867,488 different headlines and 396,001,930 positive relationships between headlines. 
}

\paragraph{Does the dataset contain all possible instances or is it a sample (not necessarily random) of instances from a larger set?}{ If the dataset is a sample, then what is the larger set? Is the sample representative of the larger set (e.g., geographic coverage)? If so, please describe how this representativeness was validated/verified. If it is not representative of the larger set, please describe why not (e.g., to cover a more diverse range of instances, because instances were withheld or unavailable).}

\textit{Many local newspapers were not preserved, and newspapers with the widest circulation tended to renew their copyrights, so cannot be included.
}

\paragraph{What data does each instance consist of? “Raw” data (e.g., unprocessed text or images) or features?}{In either case, please provide a description.}

\textit{Each data instance consists of raw data. Specifically, an example from \dataset is:} 

\begin{verbatim}
{
    "headline": "FRENCH AND BRITISH BATTLESHIPS IN MEXICAN WATERS",
    "group_id": 4
    "date": "May-14-1920",
    "state": "kansas",
}
\end{verbatim}

\textit{The data fields are: 
}

 - \verb|headline|\textit{: headline text.}
 
 - \verb|date|\textit{: the date of publication of the newspaper article, as a string in the form mmm-DD-YYYY.}
 
 - \verb|state|\textit{: state of the newspaper that published the headline.}
 
 - \verb|group_id|\textit{: a number that is shared with all other headlines for the same article. This number is unique across all year files. }

\paragraph{Is there a label or target associated with each instance?}{If so, please provide a description.}

\textit{Each instance contains a} \verb|group_id| \textit{as mentioned directly above. This is a number that is shared by all other instances that are positive semantic similarity pairs.}

\paragraph{Is any information missing from individual instances?}{If so, please provide a description, explaining why this information is missing (e.g., because it was unavailable). This does not include intentionally removed information, but might include, e.g., redacted text.}

\textit{In some cases, the state of publication is missing, due to incomplete metadata.}

\paragraph{Are relationships between individual instances made explicit (e.g., users’ movie ratings, social network links)?}{If so, please describe how these relationships are made explicit.}

\textit{Relationships between instances are made explicit in the} \verb|group_id| \textit{variable, as detailed above.}

\paragraph{Are there recommended data splits (e.g., training, development/validation, testing)?}{If so, please provide a description of these splits, explaining the rationale behind them.}

\textit{There are no recommended splits.}

\paragraph{Are there any errors, sources of noise, or redundancies in the dataset?}{If so, please provide a description.}

\textit{The data is sourced from OCR'd text of historical newspapers. Therefore some of the headline texts contain OCR errors.}

\paragraph{Is the dataset self-contained, or does it link to or otherwise rely on external resources (e.g., websites, tweets, other datasets)?}{If it links to or relies on external resources, a) are there guarantees that they will exist, and remain constant, over time; b) are there official archival versions of the complete dataset (i.e., including the external resources as they existed at the time the dataset was created); c) are there any restrictions (e.g., licenses, fees) associated with any of the external resources that might apply to a future user? Please provide descriptions of all external resources and any restrictions associated with them, as well as links or other access points, as appropriate.}

\textit{The data is self-contained.}

\paragraph{Does the dataset contain data that might be considered confidential (e.g., data that is protected by legal privilege or by doctor-patient confidentiality, data that includes the content of individuals non-public communications)?}{If so, please provide a description.}

\textit{The dataset does not contain information that might be viewed as confidential.}

\paragraph{Does the dataset contain data that, if viewed directly, might be offensive, insulting, threatening, or might otherwise cause anxiety?}{If so, please describe why.}

\textit{The headlines in the dataset reflect diverse attitudes and values from the period in which they were written, 1920-1989, and contain content that may be considered offensive for a variety of reasons.}

\paragraph{Does the dataset relate to people?}{If not, you may skip the remaining questions in this section.}


\textit{Many news articles are about people.}

\paragraph{Does the dataset identify any subpopulations (e.g., by age, gender)?}{If so, please describe how these subpopulations are identified and provide a description of their respective distributions within the dataset.}

\textit{The dataset does not specifically identify any subpopulations.}

\paragraph{Is it possible to identify individuals (i.e., one or more natural persons), either directly or indirectly (i.e., in combination with other data) from the dataset?}{If so, please describe how.}

\textit{If an individual appeared in the news during this period, then headline text may contain their name, age, and information about their actions.}

\paragraph{Does the dataset contain data that might be considered sensitive in any way (e.g., data that reveals racial or ethnic origins, sexual orientations, religious beliefs, political opinions or union memberships, or locations; financial or health data; biometric or genetic data; forms of government identification, such as social security numbers; criminal history)?}{If so, please provide a description.}

\textit{All information that it contains is already publicly available in the newspapers used to create the headline pairs.}

\paragraph{Any other comments?} ~\\
\textit{
None.
}

\bigskip
\subsubsection{Collection Process}

\paragraph{How was the data associated with each instance acquired?}{Was the data directly observable (e.g., raw text, movie ratings), reported by subjects (e.g., survey responses), or indirectly inferred/derived from other data (e.g., part-of-speech tags, model-based guesses for age or language)? If data was reported by subjects or indirectly inferred/derived from other data, was the data validated/verified? If so, please describe how.}

\textit{To create \dataset, we digitized front pages from off-copyright newspapers spanning 1920-1989. Historically, around half of articles in U.S. local newspapers came from newswires like the Associated Press. While local papers reproduced articles from the newswire, they wrote their own headlines, which form abstractive summaries of the associated articles. We associate articles and their headlines by exploiting document layouts and language understanding. We then use deep neural methods to detect which articles are from the same underlying source, in the presence of substantial noise and abridgement. The headlines of reproduced articles form positive semantic similarity pairs.}

\paragraph{What mechanisms or procedures were used to collect the data (e.g., hardware apparatus or sensor, manual human curation, software program, software API)?}{How were these mechanisms or procedures validated?}

\textit{These methods are described in detail in the main text and supplementary materials of this paper.}

\paragraph{If the dataset is a sample from a larger set, what was the sampling strategy (e.g., deterministic, probabilistic with specific sampling probabilities)?}  ~\\
\textit{The dataset was not sampled from a larger set.}

\paragraph{Who was involved in the data collection process (e.g., students, crowdworkers, contractors) and how were they compensated (e.g., how much were crowdworkers paid)?}  ~\\
\textit{We used student annotators to create the validation sets for associating bounding boxes, and the training and validation sets for clustering duplicated articles. They were paid \$15 per hour, a rate set by a Harvard economics department program providing research assistantships for undergraduates.
}

\paragraph{Over what timeframe was the data collected? Does this timeframe match the creation timeframe of the data associated with the instances (e.g., recent crawl of old news articles)?}{If not, please describe the timeframe in which the data associated with the instances was created.}

\textit{The headlines were written between 1920 and 1989. Semantic similarity pairs were computed in 2023.
}

\paragraph{Were any ethical review processes conducted (e.g., by an institutional review board)?}{If so, please provide a description of these review processes, including the outcomes, as well as a link or other access point to any supporting documentation.}

\textit{No, this dataset uses entirely public information and hence does not fall under the domain of Harvard's institutional review board.}

\paragraph{Does the dataset relate to people?}{If not, you may skip the remaining questions in this section.}

\textit{Historical newspapers contain a variety of information about people.}

\paragraph{Did you collect the data from the individuals in question directly, or obtain it via third parties or other sources (e.g., websites)?}  ~\\
\textit{The data were obtained from off-copyright historical newspapers.}

\paragraph{Were the individuals in question notified about the data collection?}{If so, please describe (or show with screenshots or other information) how notice was provided, and provide a link or other access point to, or otherwise reproduce, the exact language of the notification itself.}

\textit{Individuals were not notified; the data came from publicly available newspapers.}

\paragraph{Did the individuals in question consent to the collection and use of their data?}{If so, please describe (or show with screenshots or other information) how consent was requested and provided, and provide a link or other access point to, or otherwise reproduce, the exact language to which the individuals consented.}

\textit{The dataset was created from publicly available historical newspapers.}

\paragraph{If consent was obtained, were the consenting individuals provided with a mechanism to revoke their consent in the future or for certain uses?}{If so, please provide a description, as well as a link or other access point to the mechanism (if appropriate).}

\textit{Not applicable.}

\paragraph{Has an analysis of the potential impact of the dataset and its use on data subjects (e.g., a data protection impact analysis) been conducted?}{If so, please provide a description of this analysis, including the outcomes, as well as a link or other access point to any supporting documentation.}

\textit{No.}

\paragraph{Any other comments?} ~\\
\textit{
None.
}

\bigskip
\subsubsection{Preprocessing/cleaning/labeling}

\paragraph{Was any preprocessing/cleaning/labeling of the data done (e.g., discretization or bucketing, tokenization, part-of-speech tagging, SIFT feature extraction, removal of instances, processing of missing values)?}{If so, please provide a description. If not, you may skip the remainder of the questions in this section.}

\textit{See the description in the main text.}

\paragraph{Was the “raw” data saved in addition to the preprocessed/cleaned/labeled data (e.g., to support unanticipated future uses)?}{If so, please provide a link or other access point to the “raw” data.}

\textit{No.}

\paragraph{Is the software used to preprocess/clean/label the instances available?}{If so, please provide a link or other access point.}

\textit{No specific software was used to clean the instances.}

\paragraph{Any other comments?} ~\\
\textit{
None.
}

\bigskip
\subsubsection{Uses}

\paragraph{Has the dataset been used for any tasks already?}{If so, please provide a description.}

\textit{
No.
}

\paragraph{Is there a repository that links to any or all papers or systems that use the dataset?}{If so, please provide a link or other access point.}

\textit{
No.
}

\paragraph{What (other) tasks could the dataset be used for?}  ~\\
\textit{The dataset can be used for training models for semantic similarity, studying language change over time and studying difference in language across space.}

\paragraph{Is there anything about the composition of the dataset or the way it was collected and preprocessed/cleaned/labeled that might impact future uses?}{For example, is there anything that a future user might need to know to avoid uses that could result in unfair treatment of individuals or groups (e.g., stereotyping, quality of service issues) or other undesirable harms (e.g., financial harms, legal risks) If so, please provide a description. Is there anything a future user could do to mitigate these undesirable harms?}

\textit{The dataset contains historical news headlines, which will reflect current affairs and events of the time period in which they were created, 1920-1989, as well as the biases of this period.}

\paragraph{Are there tasks for which the dataset should not be used?}{If so, please provide a description.}

\textit{It is intended for training semantic similarity models and studying semantic variation across space and time.}

\paragraph{Any other comments?}  ~\\
\textit{
None
}

\bigskip
\subsubsection{Distribution}

\paragraph{Will the dataset be distributed to third parties outside of the entity (e.g., company, institution, organization) on behalf of which the dataset was created?}{If so, please provide a description.}

\textit{Yes. The dataset is available for public use.}

\paragraph{How will the dataset will be distributed (e.g., tarball on website, API, GitHub)}{Does the dataset have a digital object identifier (DOI)?}

\textit{The dataset is hosted on huggingface. Its DOI is 10.57967/hf/0751.}

\paragraph{When will the dataset be distributed?}  ~\\
\textit{The dataset was distributed on 7th June 2023.}

\paragraph{Will the dataset be distributed under a copyright or other intellectual property (IP) license, and/or under applicable terms of use (ToU)?}{If so, please describe this license and/or ToU, and provide a link or other access point to, or otherwise reproduce, any relevant licensing terms or ToU, as well as any fees associated with these restrictions.}

\textit{The dataset is distributed under a Creative Commons CC-BY license. The terms of this license can be viewed at \url{https://creativecommons.org/licenses/by/2.0/}}

\paragraph{Have any third parties imposed IP-based or other restrictions on the data associated with the instances?}{If so, please describe these restrictions, and provide a link or other access point to, or otherwise reproduce, any relevant licensing terms, as well as any fees associated with these restrictions.}

\textit{There are no third party IP-based or other restrictions on the data.}

\paragraph{Do any export controls or other regulatory restrictions apply to the dataset or to individual instances?}{If so, please describe these restrictions, and provide a link or other access point to, or otherwise reproduce, any supporting documentation.}

\textit{No export controls or other regulatory restrictions apply to the dataset or to individual instances.}

\paragraph{Any other comments?}  ~\\
\textit{
None.
}

\bigskip
\subsubsection{Maintenance}

\paragraph{Who will be supporting/hosting/maintaining the dataset?}  ~\\

\textit{The dataset is hosted on huggingface.}

\paragraph{How can the owner/curator/manager of the dataset be contacted (e.g., email address)?}  ~\\

\textit{The recommended method of contact is using the huggingface `community' capacity. Additionally, Melissa Dell can be contacted at melissadell@fas.harvard.edu.}

\paragraph{Is there an erratum?}{If so, please provide a link or other access point.}

\textit{There is no erratum.}

\paragraph{Will the dataset be updated (e.g., to correct labeling errors, add new instances, delete instances)?}{If so, please describe how often, by whom, and how updates will be communicated to users (e.g., mailing list, GitHub)?}

\textit{We have no plans to update the dataset. If we do, we will notify users via the huggingface Dataset Card.}

\paragraph{If the dataset relates to people, are there applicable limits on the retention of the data associated with the instances (e.g., were individuals in question told that their data would be retained for a fixed period of time and then deleted)?}{If so, please describe these limits and explain how they will be enforced.}

\textit{There are no applicable limits on the retention of data.}

\paragraph{Will older versions of the dataset continue to be supported/hosted/maintained?}{If so, please describe how. If not, please describe how its obsolescence will be communicated to users.}

\textit{We have no plans to update the dataset. If we do, older versions of the dataset will not continue to be hosted. We will notify users via the huggingface Dataset Card.}

\paragraph{If others want to extend/augment/build on/contribute to the dataset, is there a mechanism for them to do so?}{If so, please provide a description. Will these contributions be validated/verified? If so, please describe how. If not, why not? Is there a process for communicating/distributing these contributions to other users? If so, please provide a description.}

\textit{Others can contribute to the dataset using the huggingface `community' capacity. This allows for anyone to ask questions, make comments and submit pull requests. We will validate these pull requests. A record of public contributions will be maintained on huggingface, allowing communication to other users.}

\paragraph{Any other comments?}  ~\\
\textit{
None.
}

\clearpage 
 
\bibliographystyle{acm}
\bibliography{cites}

\end{document}